# Unlabeled Data Deployment for Classification of Diabetic Retinopathy Images Using Knowledge Transfer

*Sajjad Abbasi[1], Mohsen Hajabdollahi[1], Nader Karimi[1], Shadrokh Samavi[1,2], Shahram Shirani[2]*
[1]Department of Electrical and Computer Engineering, Isfahan University of Technology, Isfahan, 84156-83111 Iran
[2]Department of Electrical and Computer Engineering, McMaster University, Hamilton, ON L8S 4L8, Canada

**ABSTRACT**

Convolutional neural networks (CNNs) are extensively beneficial for medical image processing. Medical images are plentiful, but there is a lack of annotated data. Transfer learning is used to solve the problem of lack of labeled data and grants CNNs better training capability. Transfer learning can be used in many different medical applications; however, the model under transfer should have the same size as the original network. Knowledge distillation is recently proposed to transfer the knowledge of a model to another one and can be useful to cover the shortcomings of transfer learning. But some parts of the knowledge may not be distilled by knowledge distillation. In this paper, a novel knowledge distillation using transfer learning is proposed to transfer the whole knowledge of a model to another one. The proposed method can be beneficial and practical for medical image analysis in which a small number of labeled data are available. The proposed process is tested for diabetic retinopathy classification. Simulation results demonstrate that using the proposed method, knowledge of an extensive network can be transferred to a smaller model.

## 1. Introduction

Medical image processing is widely being implemented by convolutional neural networks (CNNs) due to their strong capability in the image features extraction and classification [1]–[3]. CNNs necessitate a large number of training samples as well as their corresponding labels. Unfortunately, the annotated data are rare, especially in the case of medical images. The unavailability of labeled medical images is due to the patient's privacy and security considerations. Transfer learning is introduced to transfer pre-trained model parameters to a model that has problems such as lack of large training data [4]–[6]. In the transfer learning, the model under transfer should have a similar structure to the base model. Therefore, there are some limitations in the use of transfer learning, which restricts the use of a predefined model.

In 2015, knowledge distillation (KD) was introduced to transfer the knowledge of a model to another one [7]. KD can be used between models with different structures to transfer the knowledge of a complex model (called the teacher) to a simple model (called the student) [8]. Although knowledge distillation has interesting applications, a question is aroused about the comparison of KD and transfer learning. The question is whether it is possible to use KD as an alternative to the transfer-learning or not. To answer this question, we investigate the application of knowledge distillation as an alternative for the transfer-learning. This study is aiming to design a method which has two characteristics. The first one is an appropriate knowledge transfer from a base network to another network (network under transfer). The second one is designing the model under transfer, which has an arbitrary structure. It is possible to create a simple model by using the proposed method, which utilizes an appropriate transferred knowledge. This method has interesting applications for medical images which suffer from a limited amount of annotated data. We test our approach for the classification of diabetic retinopathy (DR) images. DR classification is a challenging task in which there are not enough training data.

The remainder of this paper is structured as follows. Previous studies in DR classification are briefly described in Section 2. In Section 3, the proposed method for knowledge transfer using an unlabeled dataset is explained. In Section 4, experimental results are presented. Finally, Section 5 is dedicated to the conclusion of this study.

## 2. Diabetic Retinopathy Classification

Diabetes is a common disease that could harm the microvessels in the human eye retina [9], [10]. The advanced stage of this disease can lead to diabetic retinopathy (DR), which is considered a prevalent cause of vision loss. Regular retinal monitoring by an expert can be used to prevent vision loss, which is difficult due to its cost and lack of expert accessibility.

Automatic screening and analysis of the retina can be considered as a solution to this problem. Among different proposed methods and techniques for automatic screening of the retina, the use of CNNs can be regarded as one of the best approaches. Different networks are proposed in the literature, which their structures are very complex. For example, in [11-13], multiple network structures are utilized, which work either parallel or sequentially. Each network could have a part of the image as its input. In [2], [14], and [15], VGG based networks are proposed for DR classification. In [2], and [15], VGG network parameters are enhanced using transfer learning from a VGG model pre-trained on Image-Net dataset

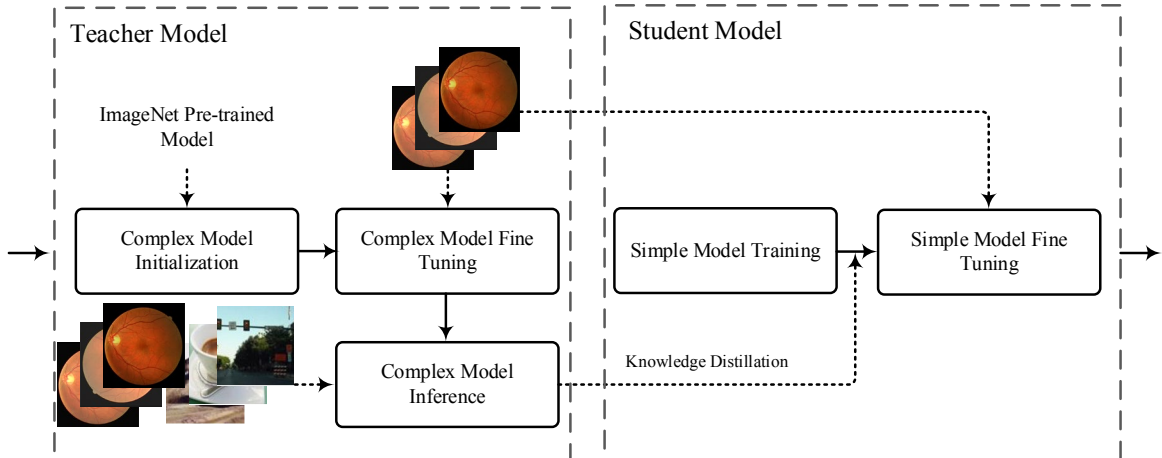
Fig. 1. Block diagram of the proposed method.

[16]. Since pre-trained structures are available in the form of a VGG network, [2], and [15] used a VGG. By reviewing different CNN structures that are used for DR analysis, it can be seen that complex networks are employed.

## 3. Proposed Method

The proposed method is based on three techniques, including transfer learning, knowledge distillation, and employing unlabeled data. In Fig. 1, the general pipeline of the proposed method is illustrated, which contains two main parts including teacher modeling and student modeling. A teacher model is a temporal model used to train the final student model, and at the inference time, only the student model is active. In the following, the proposed methods are explained.

### 3.1. Preprocessing

Before training the network structure, applying preprocessing and augmentation can be useful for better training. For preprocessing the same method as performed in [2] is utilized. Histogram equalization of the retinal images increases the contrast of the vessels, especially microvessels, and better represents abnormal regions for DR classification. For preprocessing, local histogram equalization is performed separately on each input channel. For augmentation, the equalized image, row-wise, and column-wise flipping are used to increase our training set.

### 3.2. Teacher model

In the teacher modeling stage, a complex model is trained for the DR classification. For training the teacher model, for example, a VGG structure is considered as the teacher. VGG is selected because its pre-trained version on ImageNet is available. At first, a pre-trained VGG model, trained on Image-Net, is used to initialize the teacher model. After that, the target augmented dataset is fed to the teacher model, and the teacher model is trained on the target dataset. In this way, a network with general feature extraction capability specialized on the target dataset is resulted. In this stage, the network structure is ready for knowledge distillation. Thanks to the distillation process, it is possible to train a model in which its structure is different from the teacher model. However, the question is whether it is possible to transfer all the knowledge of a teacher to a student through the distillation.

A simple and intuitive experiment is set up to answer the above ambiguity. In this experiment, a VGG network as the teacher and another VGG network as a student model are considered. The teacher model is initialized by a VGG model, which is pre-trained on the Image-Net dataset, and it is fine-tuned using a retina dataset aiming to classify them for DR levels. The student model is trained using the distilled knowledge from the teacher model. Also, for better comparison, a VGG model is trained directly on the same retina dataset for classification of DR levels. The results of these networks are illustrated in Fig. 2. It can be observed that the VGG model with distillation has slightly better accuracy than the VGG model without any distillation. But the student model has a lower accuracy far from the teacher's accuracy.

This observation implies that all of the knowledge of a network may not be transferred through distillation. Explicitly, it can be stated that the knowledge which is transferred to the teacher model is not transferred to the student model through distillation. Although the accuracy of a simple model can be improved using knowledge distillation [17], [18], also knowledge distillation works better in conditions with limited training data as stated in [19], [20], but in this case, knowledge is not transferred even to a model with the same size of the teacher model. To explain this problem, it can be said that during knowledge distillation, the student only sees the data in the target dataset and is trained based on the corresponding teacher soft labels. The knowledge that is related to a pre-trained model, and is being used by the transfer-learning, can be extracted by seeing the images related to that pre-trained model.

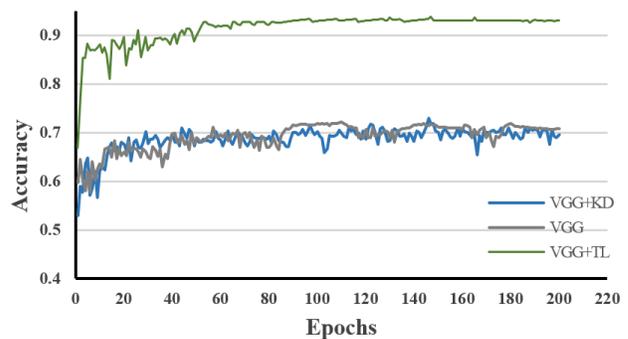

Fig. 2. Comparison of different training of VGG model for DR classification. VGG with transfer learning: VGG + TL, VGG with knowledge distillation: VGG + KD.

Now let us look at the teacher model, which is ready for the transfer-learning. In the proposed method, better knowledge transfer is predicted by using not only the labels of the retina images but also labels of random images. Random images are unlabeled, which are labeled by the teacher model. In this way, the student model extracts the knowledge of the teacher model in other images. As illustrated in Fig. 1, after fine-tuning the teacher model, knowledge for distillation are provided from both of the DR and random images.

### 3.3. Student model

In the proposed method, we are going to train a simple model as a student such that maximum information from a teacher model could be utilized. Also, training a simple model directly by a massive dataset, such as ImageNet, is a formidable task. Therefore, using unlabeled data could be an alternative to training on a large data set.

In this regard, the student model is selected as a simple model that employs the knowledge of the teacher as much as possible. As illustrated in Fig. 1, after teacher training, the student is trained in two steps. The student follows the same training trend as conducted in the teacher model. In the first step, the student is trained based on the knowledge of the random images using knowledge distillation, which simulates the transfer learning of the teacher model. After that, retina images are used to fine-tune the student model using soft labels resulted from the teacher model. This stage also simulates the fine-tuning of the teacher model on the retina images. Finally, the student can be fine-tuned again using hard labels of the DR images. In this way, the student beside training on the DR images trains on the other knowledge, which is embedded in the teacher model.

### 3.4. Knowledge distillation formulation

The way that the knowledge is distilled has an important role in better transfer of the knowledge from the teacher to the student. In [21], a teacher-student model complete which predictions of the teacher are compared with the original labels. If the prediction is correct, soft labels are used for distillation. In the proposed method, we use conditional distillation. Suppose that we have a teacher $T$ with parameters $w_T$ and a student model $S$ with parameters $w_S$. A set of training sample $D = \{d_1, d_2, ..., d_N\}$, and corresponding labels $L = \{l_1, l_2, ..., l_N\}$ with $(l_i \in \mathbb{R}^{|C|})$ on DR classification as a target dataset is considered and C is set of all possible classes of $l_i$. Also a set of random images $R = \{r_1, r_2, ..., r_M\}$ without any labels are considered. Two losses can be defined based on what stated before. The first loss is due to the unlabeled data, which can be formulated as equation (1).

$$L(w_S)_1 = \frac{-1}{M} \sum_{i=1}^{M} \sum_{j=1}^{C} p(r_i: j | T: w_T) \times \log(p(r_i: j | S: w_S)) \quad (1)$$

where $p$ is the probability, T and S represent the teacher and student models, respectively. A second loss also can be defined due to the labeled data, which is conditional as equation (2). The first term of summation indicates the loss due to the samples in which the teacher correctly predicts their labels. The second term indicates the loss of the samples, which are not correctly predicted by the teacher. In equation (2), $Ind(x)$ represents an indicator function, which is 1 when $x$ is true and 0 when $x$ is false. At the first stage of student model training, the student is trained based on the $L(w_S)_1$ until reach to an acceptable accuracy. At the second step, the student is fine-tuned using $L(w_S)_2$.

$$L(w_S)_2 = \frac{-1}{N} \sum_{i=1}^{N} \left[ Ind\left(\underset{c \in C}{Argmax}(p(d_i: c | T: w_T)) == l_i\right) \times \left( \sum_{j=1}^{|C|} p(d_i: j | T: w_T) \times \log(p(d_i: j | S: w_S)) \right) + Ind\left(\underset{c \in C}{Argmax}(p(d_i: c | T: w_T)) \neq l_i\right) \times \log(p(d_i: l_i | S: w_S)) \right] \quad (2)$$

## 4. Experimental Results

Experimental results are conducted in the case of DR classification in retina images. Messidor image dataset is used to evaluate the proposed method for DR classification [22]. In the Messidor database, there are 1200 RGB images, which we resized them to 300×300. After enhancement, by augmentation, we increased the number of images to 4800. DR classification accuracy and area under the curve of ROC (receiver operating characteristic curve) are used to evaluate the performance of different structures using five-fold cross-validation method. For classification, we follow the same definition of DR grading levels, as mentioned in [2] and [15]. All of the models are implemented by Python using the TensorFlow framework. A computer with an Nvidia GPU1080 Ti, and 11GB internal memory is used to train and test the models. For the unlabeled data, a set of natural images are randomly selected from the internet containing 20,000 images, which are resized to 300×300.

For the teacher model, a VGG model was employed. This model is not able to yield acceptable results. This problem can be due to the lack of data for the training and weak feature extraction capabilities that by using only DR data occurred. In [2] and [15], transfer learning is used to improve their results. By employing the transfer learning technique, it is possible to provide a better learning capability. To this aim, parameters of a VGG network which are pre-trained on ImageNet are used to initialize the teacher parameters.

For the student model, designing small network structures are under consideration. Small structures are different from the VGG network, which means that it is not possible to use a pre-trained VGG model. Also, training small models directly on the ImageNet can be a very time-consuming process with a lot of hardware resources. In this experiment, small models are enriched using transfer learning, knowledge distillation, and using unlabeled data. Two small versions of the VGG network with 16 layers are used, including VGG/4 and VGG/2, in which the number of their filters are divided by 4 and 2, respectively. Also, for better evaluation, a random and small structure with ten convolutional layers is utilized, which is called the "Simple10" network. Simple10 network has 20, 20, 30, 30, 40, 40, 160, 160, 250, 250 convolutional filters, in their layers. The conventional training on DR images is named as "Base," learning using knowledge distillation is named as "KD," and employing unlabeled data is called "UL." In Fig. 3, the results of different training methods for DR classification are illustrated. Fig.3a is related to the results of the simple10 network, and Fig 3.b is related to the VGG/2 network. It can be observed that using unlabeled data have an important effect on better training of simple networks. Using only knowledge distillation, in simple10 network, improves the network accuracy, but in the case of VGG/2, we do not observe any improvement. Using unlabeled data leads to a suitable improvement in accuracy in both experiments. Also, using knowledge distillation as well as unlabeled data, slightly better results are observed. Fig. 3 demonstrate that using unlabeled data could improve the training capability of a model. For a better comparison of different methods, three mentioned networks are trained for 300 epochs, and their final accuracies are reported in Tables 1 and 2 where the best results are bolded. It can be seen that using unlabeled data detection accuracies as well as the AUCs in all of the simple networks are improved. Significant differences are observed between the performance of basic training and training using unlabeled data. Finally, we can say that, by knowledge distillation and using unlabeled data, better knowledge of a network can be transferred from a teacher model to a student model.

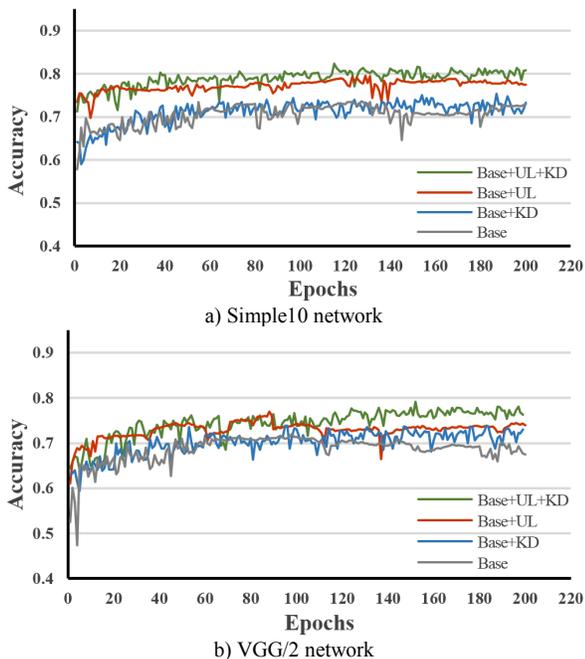

Fig. 3. Accuracies of different training methods for DR classification.

Table 1. Comparison of accuracies (percentages) of different training methods for DR classification.

| Network / Training method | VGG/4 | VGG/2 | Simple10 |
|---|---|---|---|
| Base model | 73.37 | 73.58 | 72.11 |
| Base model+KD | 76.84 | 75.37 | 73.89 |
| Base model+UL | 78.63 | 79.58 | 76.95 |
| Base model+KD+UL | **79.89** | **82.32** | **79.16** |

## 5. Conclusion

A new method for knowledge transfer from a complex network to an arbitrary simple network was proposed. The proposed algorithm employed the soft labels of a random dataset produced by a complex model to extract all of the model information. This information was used to train a simple model that was not able to perform an appropriate classification. Experimental results, for DR classification, demonstrated that the proposed use of unlabeled data by simple models improved their accuracy by an average of 5%. There are applications in medical or general image analysis that portable devices with limited resources have to be used. The proposed approach can be used as a transfer knowledge method where the available implementation platform has constraints, the design has to be simple, and the training data is limited.

Table 2. Comparison of AUCs (percentages) of different training methods for DR classification.

| Network / Training method | VGG/4 | VGG/2 | Simple10 |
|---|---|---|---|
| Base model | 78.60 | 80.18 | 78.52 |
| Base model+KD | 83.63 | 81.59 | 79.49 |
| Base model+UL | 85.17 | 87.09 | 83.34 |
| Base model+KD+UL | **86.65** | **88.92** | **85.95** |